\documentclass[conference]{IEEEtran}
\IEEEoverridecommandlockouts
\usepackage{cite}
\ifCLASSINFOpdf
   \usepackage[pdftex]{graphicx}
\else
\fi
\usepackage{amsmath}
\usepackage{amssymb}

\usepackage{array}
\ifCLASSOPTIONcompsoc
 \usepackage[caption=false,font=normalsize,labelfont=sf,textfont=sf]{subfig}
\else
 \usepackage[caption=false,font=footnotesize]{subfig}
\fi
\usepackage[linesnumbered,ruled,vlined]{algorithm2e}
\usepackage{algpseudocode}
\usepackage{graphicx}
\usepackage{xspace}
\usepackage{multicol}
\usepackage{multirow}
\usepackage{lettrine}
\usepackage{textcomp}
\usepackage{xcolor}
\def\BibTeX{{\rm B\kern-.05em{\sc i\kern-.025em b}\kern-.08em
   T\kern-.1667em\lower.7ex\hbox{E}\kern-.125emX}}

\makeatletter
\def\ps@headings{%
\def\@oddhead{\mbox{}\scriptsize\rightmark \hfil \thepage}%
\def\@evenhead{\scriptsize\thepage \hfil \leftmark\mbox{}}%
\def\@oddfoot{}%
\def\@evenfoot{}}
\makeatother
\usepackage{textcomp}
\usepackage{xcolor}
\def\BibTeX{{\rm B\kern-.05em{\sc i\kern-.025em b}\kern-.08em
    T\kern-.1667em\lower.7ex\hbox{E}\kern-.125emX}}
    
\begin{document}

%\title{Outage Analysis of an Interference Assisted Energy Harvesting with Decode-and Forward Relaying}
\title{A Comparative Analysis on Metaheuristic Algorithms Based Vision Transformer Model for Early Detection of Alzheimer's Disease
%\thanks{This research work is supported by DST, SERB, Govt. of India via sanction order no. ECR/2017/000440}
}
\author{\IEEEauthorblockN{Anuvab Sen$^1$, Udayon Sen$^2$ and Subhabrata Roy$^3$}
\IEEEauthorblockA{$^{1,3}$Department of ETCE and $^2$Department of CST\\
IIEST Shibpur, Howrah -711103, India\\
Email: sen.anuvab@gmail.com$^1$, udayon.sen3@gmail.com$^2$ and subhabrata\_ece@yahoo.com$^3$}\vspace{-1 cm}}

%\author{\IEEEauthorblockN{Subhabrata Roy}
%\IEEEauthorblockA{\textit{Dept. of IEE} \\
%\textit{Jadavpur University}\\
%Kolkata, India-700106 \\
%subhabrata\_ece@yahoo.com}
%\and
%\IEEEauthorblockN{Abhijit Chandra}
%\IEEEauthorblockA{\textit{Dept. of IEE} \\
%\textit{Jadavpur University}\\
%Kolkata, India-700106 \\
%abhijit922@yahoo.co.in}
%}




\maketitle 

\begin{abstract}
A number of life threatening neuro-degenerative disorders had degraded the quality of life for the older generation in particular. Dementia is one such symptom which may lead to a severe condition called Alzheimer's disease if not detected at an early stage. It has been reported that the progression of such disease from a normal stage is due to the change in several parameters inside the human brain. In this paper, an innovative metaheuristic algorithms based ViT model has been proposed for the identification of dementia at different stage. A sizeable number of test data have been utilized for the validation of the proposed scheme. It has also been demonstrated that our model exhibits superior performance in terms of accuracy, precision, recall as well as F1-score.
\end{abstract}

% \fi
\IEEEpeerreviewmaketitle

\vspace{-.2 cm}

\section{Introduction}
Dementia is considered to be one of the rapidly growing neurological disorders mostly among the older population in past few years that leads to short term memory loss, disorganized cognitive and motor action, lack of recognition and eventually results in death \cite{gauthier2021world}. Dementia, when remains untreated at an early stage, results in a specific neuro-psychiatric disorder called Alzheimer's disease (AD) \cite{roy2022detection}. At present, over 5 crore individuals worldwide are grappling with Alzheimer's disease (AD), and India alone accounts for more than 6 million cases. Till now, no specific medicines are available in the world for the treatment of this disease and hence a number of measures are taken to restrict its further progression.

Over the last 20 years, different approaches have been adopted by various researchers in classifying the patient data into appropriate categories by extracting meaningful features from brain MRI images. This involves voxel-based analysis, ROI based approach, machine learning tools, neural network models and its variants, patch-based approach, statistical analysis etc. However, each of these techniques suffers from significant computational burden which needs to be curbed for big data analytics. Moreover, convolutional neural network (CNN) primarily focuses on the kernel wise local computation of the input images and thereby ignoring the correlation between the part and whole images. Irrespective of these, there are other issues related to under-fitting and over-fitting in several machine learning models.

To address these challenges and inspired by the transformative impact in natural language processing (NLP) \cite{vaswani2017attention}, scientists have turned their attention to a groundbreaking architecture known as the vision transformer (ViT). This innovative approach has garnered significant interest for overcoming the limitations of convolutional neural networks (CNNs) by employing a multi-headed self-attention-based architecture. This design effectively captures long-range dependencies, enabling the model to attend to all elements in the input sequence and achieve superior performance \cite{dosovitskiy2020image}. Although ViT shows its superiority in the field of computer vision, most of the ViT-based works are based on the ImageNet dataset \cite{deng2009imagenet} only, which is a benchmark of the natural image dataset. However, for medical image analysis, particularly for the detection of AD from brain MRI images, this approach rarely finds its application. The intricacies of the human brain's highly complex network, where distant regions exhibit strong dependencies, make ViT's self-attention mechanism particularly advantageous over CNNs.

To this aim, present study explores the potential of ViT model on the medical image classification and detection task. Similar to that of the deep learning model, its performance also depends on the proper selection of the hyper-parameters. In this work, we have utilized the concept of different metaheuristic algorithms such as Differential Evolution (DE) \cite{storn1997differential}, Genetic Algorithm (GA) \cite{holland1984genetic}, Particle Swarm Optimization (PSO) \cite{kennedy1995particle}, Ant Colony Optimization (ACO) \cite{mazumder2023benchmarking} in order to obtain the best hyper-parameters. It has already been proved that the metaheuristics are more efficient, robust and scalable than other hyper-parameter search techniques such as grid search \cite{lavalle2004relationship}, random search \cite{bergstra2012random} and Bayesian Optimization \cite{snoek2012practical}. Furthermore, these algorithms can be employed to various non-linear, non-convex and non-continuous functions \cite{cao2023performance}. In a nutshell, the major contributions of the proposed work can be outlined as follows:
\begin{itemize}
\item
Introducing a vision transformer-based approach for AD detection, employing the concept of various metaheuristics algorithms for hyper-parameter selection and classifying patient data into AD, MCI, and HC classes.  
\item
3D brain MRI images have been preprocessed by utilizing the statistical parametric mapping (SPM12) tool in order to extract the 2D MRI images so that proposed ViT model can work with 3D data.
\item
Present study is unique in terms of investigating the transformer's self-attention based mechanism outside of the usual scope of natural language processing.
\item
Presenting comprehensive simulation results for performance evaluation, utilizing several metrics such as accuracy, precision, recall, F1-score followed by a comparative analysis with other novel methods.
\item
Finally, it can be concluded that proposed ViT model with the aid of different metaheuristic optimizers outperforms other state-of-the-art models in terms of hyper-parameter selection, speed and robustness.
\end{itemize}

To the best of our knowledge, such metaheuristic algorithm based hyper-parameter tuning scheme for the ViT model in order to diagnose AD seems new and no previous literature investigates such a large and diverse set of metaheuristic optimizers. The rest of the paper can be summarized as follows: Proposed metaheuristic algorithm based ViT model for the purpose of classification has been demonstrated in section II. Section III summarizes the results obtained followed by concluding remarks in Section IV.

\section{Proposed Metaheuristic algorithm based Vision Transformer Model}
A brief discussion on the typical Vision Transformer model and some of the most famous  metaheuristic algorithms namely Differential Evolution, Genetic Algorithm, Particle Swarm Optimization and Ant Colony Optimization is demonstrated in this section followed by the implementation of incorporating such metaheuristics with the ViT model for classification of AD, MCI and HC.

\subsection{Vision Transformer}
In the typical architecture of a Vision Transformer (ViT), the Transformer encoder \cite{vaswani2017attention} assumes a pivotal role, comprising multiple identical layers, each containing two sub-layers: multi-head self-attention (MSA) and multi-layer perception (MLP). A crucial aspect involves the utilization of a residual connection \cite{he2016deep} around each of these sub-layers, followed by layer normalization (LN) \cite{ba2016layer}. The input ($\mathbf{Z}_{0}$) is an array of N embedded image patches ($\mathbf{T}_{p}^{i}$) and a distinctive classification token named $\mathbf{T}_{\text{cls}}$. State of special classification token at transformer encoder's output ($\mathbf{Z}_{\mathcal{L}}^{0}$) serves as image representation ($\mathbf{Y}$) for classification task. A classification head, implemented using an MLP with a single hidden layer during pre-training and a single linear layer during fine-tuning, is connected to $\mathbf{Z}_{\mathcal{L}}^{0}$. Additionally, learnable position embeddings are introduced to the patch embeddings, and, along with $\mathbf{T}_{\text{cls}}$, they constitute the encoder's input. The step-by-step process of the vision transformer model can be succinctly summarized using the following equations:
\begin{gather}
\mathbf{Z}_{0} = [\mathbf{T}_{\text{cls}}; \mathbf{T}_{p}^{1}; \mathbf{T}_{p}^{2}; \ldots; \mathbf{T}_{p}^{N}] + \mathbf{T}_{\text{pos}}, \nonumber \\
\mathbf{T}_{\text{cls}}, \mathbf{T}_{p}^{i} \in \mathbb{R}^{1 \times D}, \quad \mathbf{T}_{\text{pos}} \in \mathbb{R}^{(N+1) \times D} \\
\mathbf{Z}_{l}^{\prime} = \text{MSA}(\text{LN}(\mathbf{Z}_{l-1})) + \mathbf{Z}_{l-1}, \quad l = 1,2, \ldots, \mathcal{L} \\
\mathbf{Z}_{l} = \text{MLP}(\text{LN}(\mathbf{Z}_{l}^{\prime})) + \mathbf{Z}_{l}^{\prime}, \quad l = 1,2, \ldots, \mathcal{L} \\
\mathbf{Y} = \text{LN}(\mathbf{Z}_{\mathcal{L}}^{0})
\end{gather}

Let \( \mathbf{Z}_{0} \) be the input with classification token \( \mathbf{T}_{\text{cls}} \) and patch embeddings \( \mathbf{T}_{p}^{i} \), plus positional embeddings \( \mathbf{T}_{pos} \). Blocks are computed iteratively as \(\mathbf{Z}_{l} = \text{MLP}(\text{LN}(\text{MSA}(\text{LN}(\mathbf{Z}_{l-1})))) + \mathbf{Z}_{l-1}\). The final output \( \mathbf{Y} \) is the layer-normalized state of the classification token after the last block.

\subsection{Differential Evolution}
Differential Evolution, first coined by Rainer Stron and Kenneth Price in the year 1997, is a simple, powerful, robust, stochastic, population-based, easy to use optimization algorithm in order to solve a wide range of objective functions which are possibly non-linear, non-differentiable, non-continuous, noisy \cite{storn1997differential}. Being an evolutionary algorithm, it always initiates with a number of D-dimensional search variable vectors. The pseudocode for the Differential Evolution is depicted in Algorithm $1$.
\RestyleAlgo{ruled}
\begin{algorithm}
    \caption{Differential Evolution}
    \label{algo:de}
    \SetAlgoNlRelativeSize{-1}

    %\KwIn{Population Size $N$, Dimension $D$, Scaling Factor $F \in (0, 2)$, Crossover Probability $CR \in [0, 1]$, Termination Criterion}
    %\KwOut{Best Individual}
    Initialize the population $P$ with $N$ random individuals in the search space\;
    \While{Termination Criterion is not met}{
        \ForEach{individual $i$ in $P$}{
            Select three distinct random individuals $a$, $b$, and $c$ from $P$\;
            Generate a trial vector $v$ by combining the components of $a$, $b$, and $c$ using the DE mutation strategy\;
            \ForEach{dimension $j$ in $D$}{
                Generate a random number $r \in [0, 1]$\;
                \If{$r < CR$ or $j$ is a random dimension}{
                    $v[j] = v[j]$\;
                }
                \Else{
                    $v[j] = i[j]$\;
                }
            }
            Evaluate the fitness $f(v)$\;
            \If{$f(v) < f(i)$}{
                Replace individual $i$ with trial vector $v$\;
            }
        }
    }
    \KwRet Best individual in the final population\;
\end{algorithm}

Differential evolution (DE) thus operates by iteratively updating the candidate solutions and evolving the population over multiple generations iteratively. The process is repeated for a specific number of generations, until a termination criterion is satisfied, or a desired level of convergence is achieved eventually.

\subsection{Particle Swarm Optimization}
Particle Swarm Optimization (PSO) is another bio-inspired metaheuristic optimization algorithm based on the behaviour of a fish or bird swarm in nature. It is one of the popular choice for solving complex optimization problems owing to its efficiency and simplicity. It aims to search and find the optimal solution in a multidimensional search space by simulating the pattern and movement of particles. Pseudocode for the Particle Swarm Optimization Algorithm is displayed in Algorithm $2$.
\begin{algorithm}
\caption{Particle Swarm Optimization (PSO)}
%\SetKwInput{Input}{Input}
%\SetKwInput{Output}{Output}

%\Input{Population Size $N$, Number of Iterations $MaxIter$, Problem-specific Initialization, Termination Criterion}
%\Output{Best Particle}

Initialize particles' positions and velocities in the search space\;
\For{$iteration = 1$ to $MaxIter$}{
    \For{each particle $i$}{
        Evaluate fitness function $f(x_i)$ for particle $i$\;
        \If{$f(x_i)$ is better than the best fitness value of particle $i$}{
            Update personal best position: $pbest[i] = x_i$\;
            Update personal best fitness value: $pbest\_value[i] = f(x_i)$\;
        }
    }
    Update global best particle: $gbest = $ particle with the best fitness among all particles\;
    \For{each particle $i$}{
        Update particle velocity and position using equations:\;
        $velocity[i] = inertia \times velocity[i] + cognitive\_coefficient \times random() \times (pbest[i] - position[i]) + social\_coefficient \times random() \times (best - position[i])$\;
        $position[i] = position[i] + velocity[i]$\;
    }
}
\Return $gbest$\;
\end{algorithm}

As evident Particle Swarm Optimization Algorithm does not use the gradient of the problem being optimized, which means PSO does not require that the optimization problem be differentiable as is required by other classic optimization methods.

\subsection{Genetic Algorithm}
Genetic Algorithm (GA) is a kind of bio-inspired metaheuristic algorithm utilizing the concept of natural selection, which is the cause of biological evolution as espoused by Darwin's theory of evolution. The way evolution rewards successful individuals in a population, the GA generates optimal solutions in a constrained environment. Pseudocode for the Genetic Algorithm is displayed in Algorithm $3$. 

Each generation cycles through the four phases until GA iterates through the maximum number of cycles or until a termination criterion is met.
\begin{algorithm}
    \caption{Genetic Algorithm}
    \SetAlgoNlRelativeSize{-1}
    
    %\KwIn{Population Size $N \in (5,10)$, Chromosome Length $L$, Termination Criterion}
    %\KwOut{Best Individual}

    Initialize the population with $N$ random individuals\;
    \While{Termination Criterion is not met}{
        \ForEach{individual $x_i$ in population}{
            Evaluate fitness $f(x_i)$ for individual $x_i$\;
        }
        Select parents $p_1$, $p_2$ from the population for mating using a Roulette Wheel Selection scheme\;
        \ForEach{selected parent pair $(p_1, p_2)$}{
            Apply uniform crossover and mutation operations on parents $p_1$ and $p_2$ to obtain child $c_i$\;
        }
        Replace the current population with the new population\;
    }

    \KwRet Best individual in the final population\;
\end{algorithm}

\subsection{Ant Colony Optimization}
Ant Colony Optimization (ACO) is a robust bio-inspired optimization algorithm based on the foraging patterns of ants in nature. It simulates the behaviour of ants in search of the most optimal path between the nest and the food source. Summarized pseudocode of the Ant Colony Swarm Optimization Algorithm is displayed in Algorithm $4$. 

\begin{algorithm}
\caption{Ant Colony Optimization}
%\SetKwInOut{Input}{Input}
%\SetKwInOut{Output}{Output}
%\Input{Number of Ants $N$, Number of Iterations $\text{MaxIter}$,
       %Pheromone Evaporation Rate $\rho$, Initial Pheromone $\tau_0$,
       %Ant Movement Control Parameters $\alpha$ and $\beta$}
%\Output{Best Solution}
Initialize pheromone levels $\tau_{ij}$ on all edges $(i, j)$ to $\tau_0$\;
Initialize bestSolution with an arbitrary solution\;
Initialize bestObjective with a large value\;
\For{$\text{iter} = 1$ \KwTo $\text{MaxIter}$}{
    \For{$\text{ant} = 1$ \KwTo $N$}{
        Initialize ant's currentSolution with an arbitrary solution\;
        \For{each step in the solution}{
            Calculate the probability of moving to each neighboring solution based on pheromone\;
            Choose the next solution using the probability distribution\;
            Update ant's currentSolution and currentObjective\;
        }
        \If{currentObjective is better than bestObjective}{
            Update bestSolution and bestObjective\;
        }
        Update pheromone levels on all edges based on ant's tour and evaporation rate $\rho$\;
    }
}
Return bestSolution\;
\end{algorithm}

It employs the population of artificial ants that traverse through a proposed solution space with the help of pheromone trails that guide future search behaviour.

\subsection{Proposed ViT Architecture}
The proposed model is built upon the premises of the Vision Transformer (ViT) architecture, which has demonstrated remarkable success in various computer vision tasks. Metaheuristic algorithms are used to maximise the fitness and find out the ideal set of hyper-parameters for the classification task and hence produces most desirable results. Data augmentation techniques were applied using TensorFlow's Sequential API which have introduced variability into the data. These techniques include resizing, random horizontal flipping, random rotation etc. Because the ViT model accepts 2D images as input, in order to adapt this model in brain imaging domain, we first preprocessed the 3D brain MRI images of one subject with the aid of Statistical Parametric Mapping (SPM12) into 2D MRI images. The ViT model begins with an input layer to receive 2D brain MRI images. Each image were resized to $224*224$ pixels. A custom embedding layer called $Patches$ is defined to extract sequence of flattened patches from input images. The size of the patches were customized to be $16*16$ pixels in size. Following the patching process each patch is tokenized and passed through a dense layer which reduces the dimensionality of each patch. The Patch-Encoder layer of the given model is designed to encode these patches, including positional embeddings.

The core of the ViT model consists of multiple Transformer encoder layers. Each layer consists of the following components:
\begin{itemize}
\item 
\textbf{Multi-Head Self-Attention Mechanism:} The self attention mechanism allows to capture global dependencies on different patches of the labelled image efficiently.
\item 
\textbf{Position-wise Feed-Forward Neural Network:} After self-attention, each patch's representation is further processed through a position-wise feed-forward neural network.
\item
\textbf{Layer Normalization and Residual Connections:} To stabilize the training procedure, layer normalization and residual connections are applied after each sub-layer.
\end{itemize}

Within each encoder layer, there is an 8-headed attention mechanism with a dimensionality of 64 and a dropout ($\mathcal{D}$) of $0.1$. It helps to simultaneously process different aspects of each of the patch sequences of an image. The dropout, applied after attention mechanism, is to enhance model generalization and hence to improve it's performance substantially. The result is then flattened into 2D tensors to treat the data as a sequence of 2D inputs which are run through multiple fully connected dense layers (i.e. MLP) \cite{dosovitskiy2020image} before returning through a 3-node output layer. Each fully connected layer consists of neurons that are connected to every neuron in the previous layer. The activation function used for all cases, except for the output layer, is Rectified Linear Units (ReLU) \cite{dosovitskiy2020image} which is defined as:
\begin{gather}
f(x) = max(0, x)
\end{gather}
For the output layer, a linear activation Function is used which is of the form like:
\begin{gather} 
f(x) = x
\end{gather}
Furthermore, the output layer consists of the softmax activation to produce the class wise probability scores. The softmax function is defined as:
\begin{gather}
softmax(z_{i}) = \frac{e^{z_{i}}}{\sum_{j=1}^{n} e^{z_{j}}}~~~~\forall~~i
\end{gather}
~~where, $z_{i}$ represents the raw score (logit) for class $i$, and $n$ indicates the total number of such classes.

The multi-headed attention is used for the model to simultaneously operate on different aspects of the image. The normalization layer is applied to make the model robust thus avoiding relying on specific features too much. This also reduces the over-fitting by a large margin. Proposed ViT architecture have made use of different metaheuristics such as DE, GA, PSO and ACO in order to optimize the hyper-parameters such as batch size ($\mathcal{B}$), learning rate ($\eta$) and epoch ($\mathcal{E}$). Suggested model is evaluated by assessing the Sparse Categorical Accuracy for each set of hyper-parameters. The metric serves as the fitness function and the optimization process aims to maximize this metric with the aid of these metaheuristic optimizers which helps the proposed model to identify the ideal set of hyper-parameters. The proposed architecture is portrayed in Fig. 1.
\begin{figure}[t]
\centering
\includegraphics[width=3.5in,height=4in]{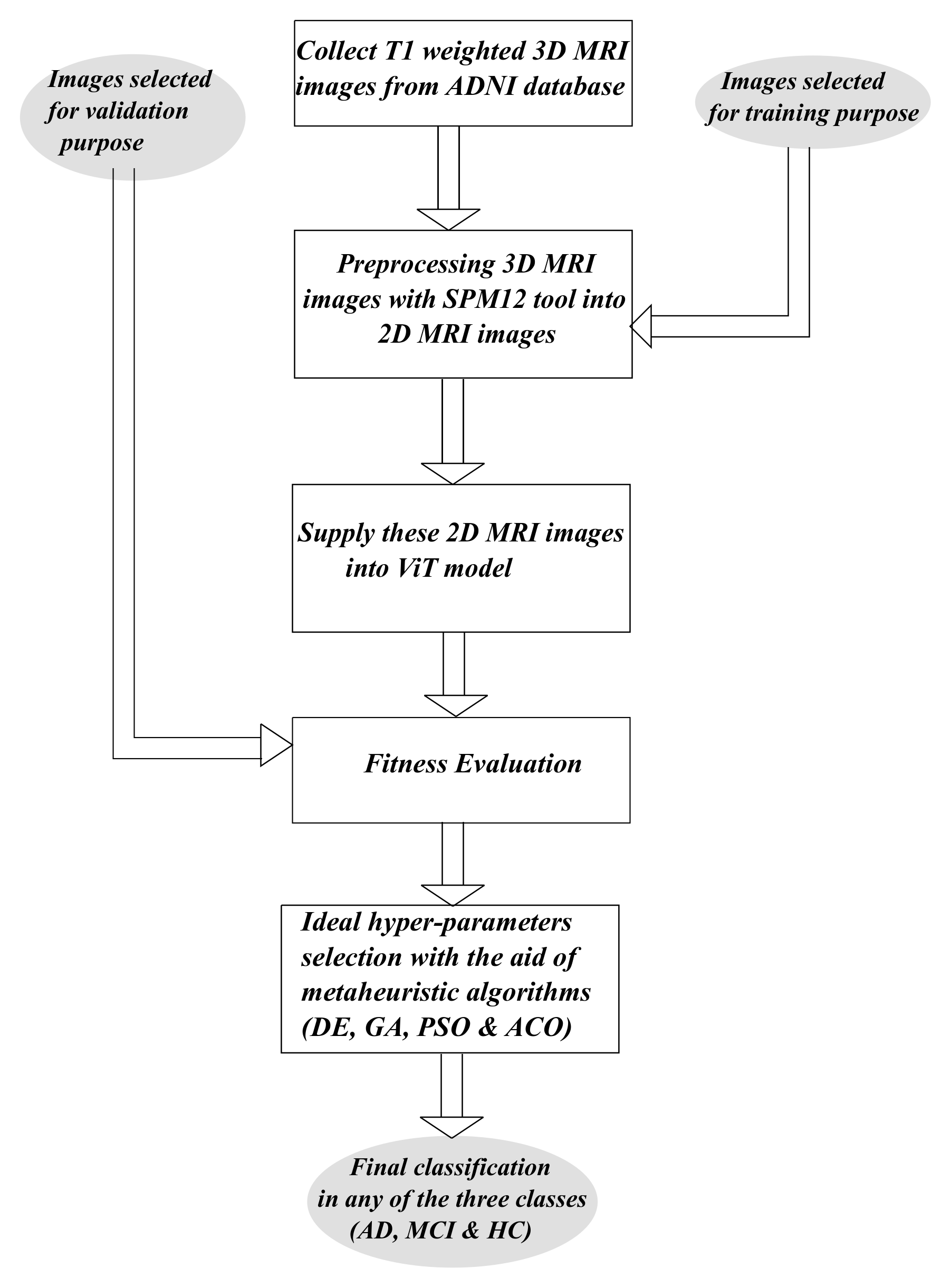}
\label{FS}
\vspace{-.18 cm}
\caption{Design flow of proposed model}
\end{figure}

\section{Experimental results}
Current investigation leverages the ADNI dataset, a globally accessible resource (http://adni.loni.usc.edu/). ADNI's overarching goal is to establish sensitive and precise approaches for early-stage AD diagnosis and to monitor AD progression through biomarkers. Our study encompasses 600 MRI scans sourced from the ADNI database, featuring diverse subject profiles, ages, series, slices, and acquisition planes. Dataset is partitioned into training (68\%), testing (20\%), followed by validation (12\%) subsets. The proposed model is trained using hyper-parameters outlined in Table I. The ADNI website provides MRI scans comprising (256 x 256 x 196) voxels, each approximately sized at $(1.0mm x 1.0mm x 1.2mm)$. MRI data, obtained in NiFTI format, underwent extraction of 2D images from 3D scans using SPM12 tool, and itk-SNAP \cite{py06nimg} served as the slice extraction tool. 

In this section, our objective is to categorize the human brain into 3 distinct classes where AD, characterized as a neurodegenerative condition, is denoted as positive (indicating the presence of the disease), while HC is treated as negative (indicating the absence of the disease). MCI, positioned as a intermediate stage in between these two classes. DE's superior performance can be achieved due to several reasons. First, it effectively explores the search space and thereby exploits promising regions for optimal solutions. Secondly, the mutation operator introduces random perturbations to prevent early convergence. Moreover, recombination facilitates the exchange of promising features, enhances the convergence speed. Finally, selection operator preserves the fittest individuals and hence enhances the quality of solutions. Given that all classification techniques are prone to the risk of misclassification, our proposed model undergoes evaluation using accuracy ($\mathcal{A}$), recall ($\mathcal{R}$), precision ($\mathcal{P}$), and F1 score. The objective is to enhance all these performance parameters simultaneously. 
\begin{table}
\caption{Ideal hyper-parameters used to train the proposed model}
\label{table_example}
\centering
\begin{tabular}
{p{2.2cm}|p{1.5cm}|p{1.5cm}} 
\hline                                                                                                                                        \multirow{2}{*}{\textbf{Hyper-parameters}} & 
\multicolumn{2}{c}{\textbf{Values}}\\
\cline{2-3}
 & DE & GA\\
\hline 
$\mathcal{B}$ & 8 & 16\\ 

$\mathcal{E}$ & 125 & 500\\ 

Input Size & 224 & 224\\ 

$\mathcal{D}$ & 0.1 & 0.1\\  

$\eta$ & 0.00067 & 0.0001\\   
\hline
\multirow{2}{*}{\textbf{Hyper-parameters}} & 
\multicolumn{2}{c}{\textbf{Values}}\\
\cline{2-3}
 & PSO & ACO\\
\hline 
$\mathcal{B}$ & 18 & 8\\ 

$\mathcal{E}$ & 238 & 148\\ 

Input Size & 224 & 224\\ 

$\mathcal{D}$ & 0.1 & 0.1\\  

$\eta$ & 0.0001 & 0.000591\\    
\hline
\end{tabular}
\end{table}
\begin{figure*}[t]
\centering
\includegraphics[width=1.8in,height=1.5in]{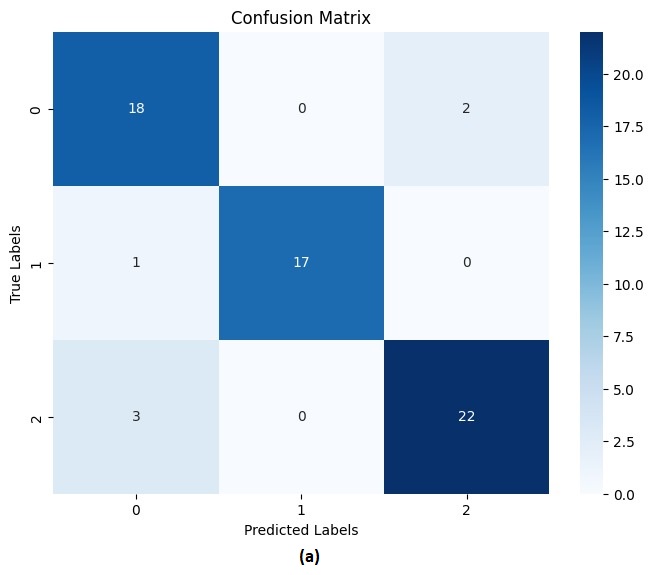}
\includegraphics[width=1.8in,height=1.5in]{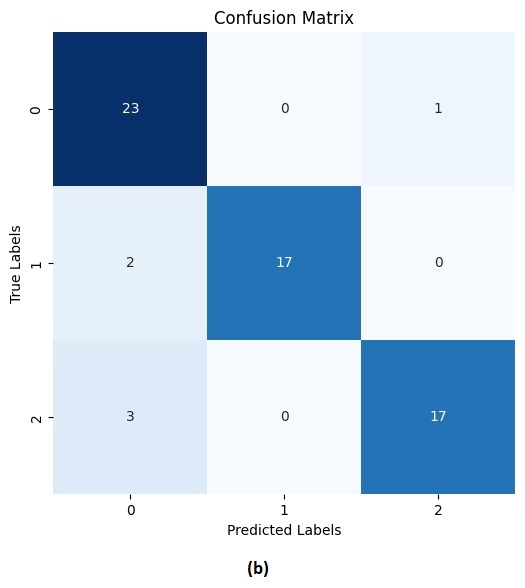}
\includegraphics[width=1.8in,height=1.5in]{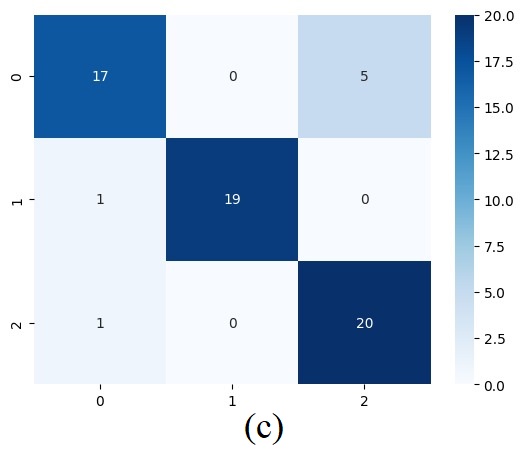}
\includegraphics[width=1.8in,height=1.5in]{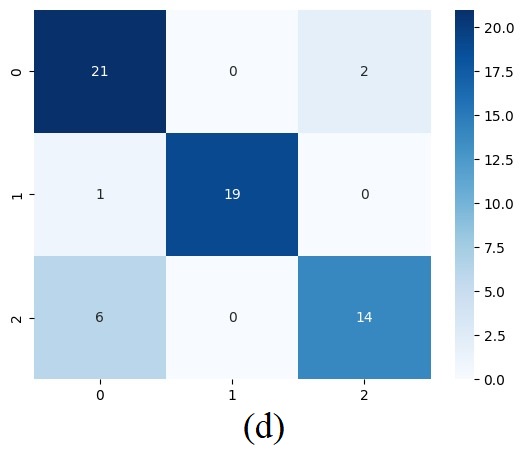}
\includegraphics[width=1.8in,height=1.5in]{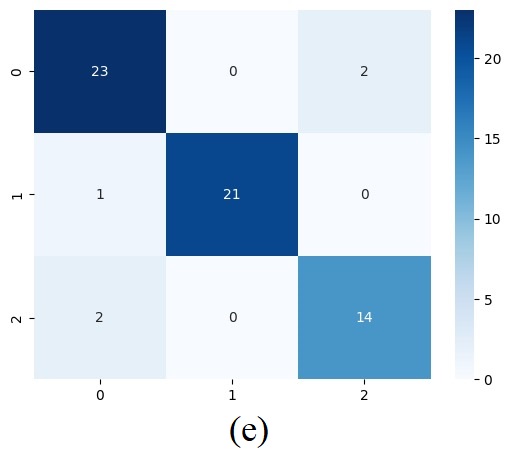}
\label{FS}
\vspace{-.18 cm}
\caption{Confusion matrix for \textbf{(a)} ViT model, \textbf{(b)} DE based ViT model, \textbf{(c)} GA based ViT model, \textbf{(d)} PSO based ViT model, \textbf{(e)} ACO based ViT model}
\end{figure*}

Fig. 2 displays the confusion matrices of traditional ViT model and proposed metaheuristic algorithms based ViT model where the accuracy of the proposed DE based ViT model is obtained as 96.8\% which is averaged over five statistical runs; whereas, other metaheuristic algorithms based ViT models achieve a classification accuracy of 91\%, 92\% and 94\% respectively with GA, PSO and ACO. However, for traditional ViT model accuracy is obtained as 92.06\%. Performance metrics, as observed with the aid of such metaheuristic algorithms based ViT model, have been compared with some of the SOTA techniques of AD detection in Table III below.  PSO and ACO may perform better than GA due to its ability to efficiently explore the search space. PSO’s inherent exploration mechanism helps it navigate the search space effectively and avoid getting stuck in local minima, unlike GA. However, it is observed that ACO's performance can be sensitive to its parameter settings such as pheromone update rules and exploitation balance. Tuning these parameters proved to be a challenge while testing.
\begin{table}[h]
\caption{Comparison of the performance metrics in terms of accuracy, recall, precision and F1-score for early detection of AD}
\label{table_example}
\centering
\begin{tabular}
{p{2.5cm}|p{1cm}|p{1cm}|p{1cm}|p{1cm}} 
\hline
Models & $\mathcal{A}$ & $\mathcal{R}$ & $\mathcal{P}$ & $F1$\\
\hline
Korolev et al. \cite{korolev2017residual} & 0.79 & 0.7305 & 0.7771 & 0.7531\\

Huang et al. \cite{huang2019diagnosis} & 0.895 & 0.8563 & 0.8994 & 0.8773\\

Shin et al. \cite{shin2023vision} & 0.8 & 0.6 & 0.75 & 0.6667\\

VGG19 \cite{shin2023vision} & 0.7333 & 0.6 & 0.6 & 0.6\\

Sherwani et al. \cite{sherwani2023comparative} & 0.902 & 0.74 & 0.92 & 0.68\\

Lyu et al. \cite{lyu2022classification} & 0.953 & 0.944 & 0.9 & -\\

Kushol et al. \cite{kushol2022addformer} & 0.882 & 0.956 & - & -\\

Hu et al. \cite{hu2023vgg} & 0.772 & 0.7997 & - & -\\

Proposed method with DE & 0.968 & 0.94 & 0.95 & 0.96\\

Proposed method with GA & 0.91 & 0.86 & 0.89 & 0.91\\

Proposed method with PSO & 0.92 & 0.84 & 0.89 & 0.88\\

Proposed method with ACO & 0.94 & 0.94 & 0.94 & 0.95\\
\hline
\end{tabular}
\end{table}

The data in Table III unmistakably demonstrate that our architecture outperforms current novel methods with respect to accuracy, recall, precision, and F1-measure. Achieving high values across all these parameters simultaneously poses a well-known challenge. Furthermore, it is important to note that our proposed ViT model, utilizing metaheuristic algorithms, stands out in contrast to conventional machine learning classification methods. This innovative approach significantly reduces computational demands, enabling effective handling of extensive MRI datasets. This distinctive feature underscores the ViT-based classifier's prowess in early Alzheimer's disease detection.

\section{Conclusion}
In this publication, we introduce a novel ViT model utilizing metaheuristics for dementia identification. Analysis of simulation outcomes reveals that our proposed model achieves superior classification performance, boasting an accuracy of approximately 96.8\%. Notably, it preserves precision, recall, and F1-score at the desired levels when compared to existing techniques. Looking ahead, there is potential for the application of our model to address additional neurological disorders, including early mild cognitive impairment (EMCI) and late mild cognitive impairment (LMCI).

\vspace{-.1 cm}
\bibliography{reference}
\end{document}